\documentclass[sigconf]{acmart}

\usepackage{makecell}

\setcopyright{rightsretained}

\copyrightyear{2025}
\acmYear{2025}
\setcopyright{othergov}
\acmConference[ICVGIP 2025]{Indian Conference on Computer Vision, Graphics, and Image Processing}{December 17--20, 2025}{Mandi, India}
\acmBooktitle{Indian Conference on Computer Vision, Graphics, and Image Processing (ICVGIP 2025), December 17--20, 2025, Mandi, India}
\acmPrice{}
\acmDOI{10.1145/3774521.3774607}
\acmISBN{979-8-4007-1930-1/2025/12}
\begin{document}
\title{Improving Video Question Answering through query-based frame selection}

\author{Himanshu Patil} 
\authornote{Equal contribution}
\affiliation{%
  \institution{Indian Institute of Technology, Bombay}
  \streetaddress{}
  \city{}
  \state{}
  \country{}
  \postcode{}
}

\author{Geo Jolly}
\authornotemark[1]
\affiliation{%
  \institution{Indian Institute of Technology, Bombay}
  \streetaddress{}
  \city{}
  \state{}
  \country{}
  \postcode{}
}

\author{Ramana Raja Buddala}
\authornotemark[1]
\affiliation{%
  \institution{Indian Institute of Technology, Bombay}
  \streetaddress{}
  \city{}
  \state{}
  \country{}
  \postcode{}
}

\author{Ganesh Ramakrishnan}
\affiliation{%
  \institution{Indian Institute of Technology, Bombay}
  \streetaddress{}
  \city{}
  \state{}
  \country{}
  \postcode{}
}

\author{Rohit Saluja}
\affiliation{%
  \institution{Indian Institute of Technology, Mandi}
  \streetaddress{}
  \city{}
  \state{}
  \country{}
  \postcode{}
}

\renewcommand{\shortauthors}{}

\begin{abstract}
Video Question Answering (VideoQA) models enhance understanding and interaction with audiovisual content, making it more accessible, searchable, and useful for a wide range of fields such as education, surveillance, entertainment, and content creation. Due to heavy compute requirements, most large visual language models (VLMs) for VideoQA rely on a fixed number of frames by uniformly sampling the video. However, this process does not pick important frames or capture the context of the video. We present a novel query-based selection of frames relevant to the questions based on the submodular mutual Information (SMI) functions. By replacing uniform frame sampling with query-based selection, our method ensures that the chosen frames provide complementary and essential visual information for accurate VideoQA. We evaluate our approach on the MVBench dataset, which spans a diverse set of multi-action video tasks. VideoQA accuracy on this dataset was assessed using two VLMs, namely Video-LLaVA and LLaVA-NeXT, both of which originally employed uniform frame sampling. Experiments were conducted using both uniform and query-based sampling strategies. An accuracy improvement of up to \textbf{4\%} was observed when using query-based frame selection over uniform sampling. Qualitative analysis further highlights that query-based selection, using SMI functions, consistently picks frames better aligned with the question. We opine that such query-based frame selection can enhance accuracy in a wide range of tasks that rely on only a subset of video frames.

\end{abstract}

%
%


\maketitle
\begin{figure*}
\centering
\includegraphics[width =\textwidth]{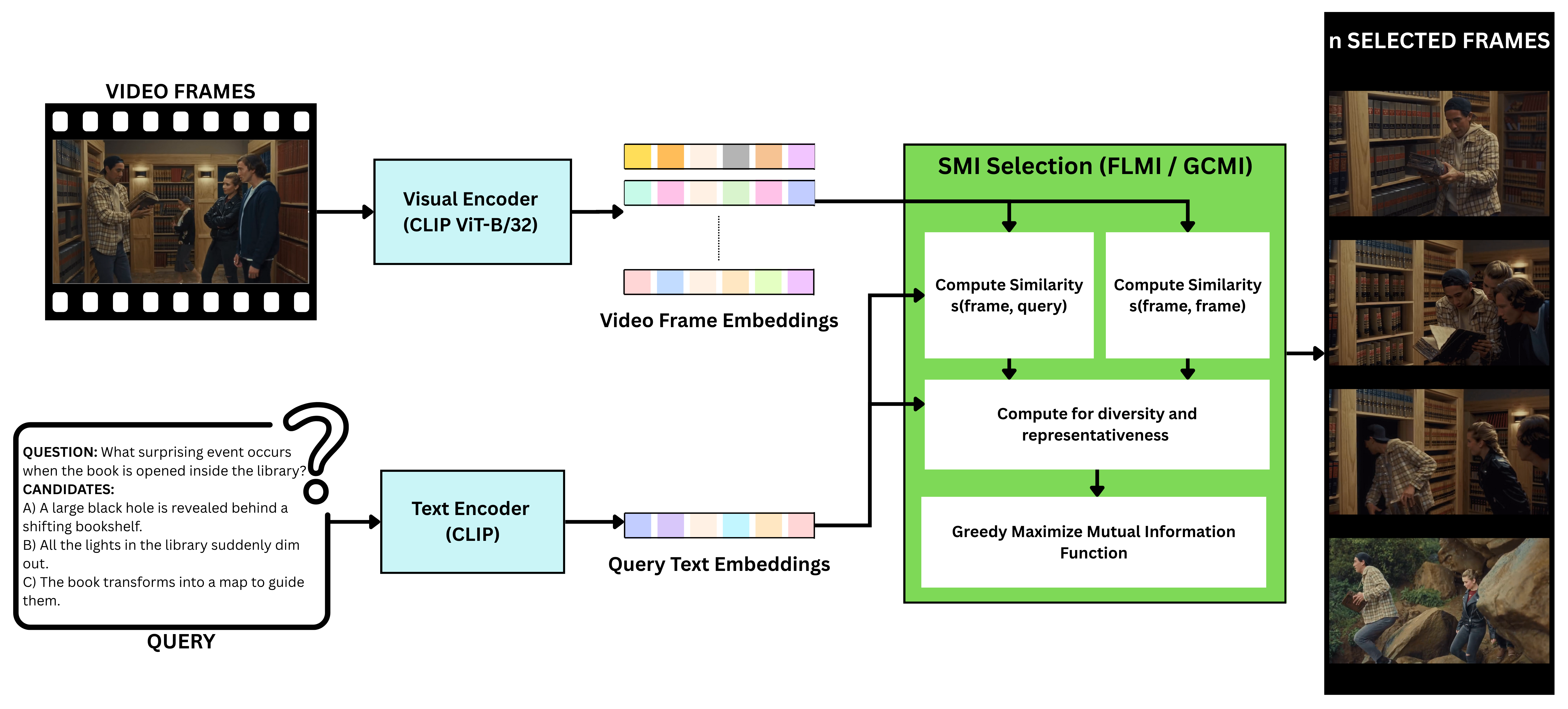}
\caption{Figure illustrates how the frames are selected employing Query based selection}
\label{fig1}
\end{figure*}

\section{Introduction}
The proliferation of video content across domains ranging from social media and entertainment to surveillance and remote education has led to an urgent need for VLMs capable of understanding and interacting with videos in an intelligent and scalable manner. VLMs interact with audio-visual content in varied ways such as Video Summarization, Video Captioning, VideoQA, and Video retrieval. Among these, VideoQA has emerged as a key task, requiring models to reason over visual and temporal information to answer natural language questions about video content. Applications of VideoQA span video analytics, human–robot interaction, accessibility tools, and intelligent tutoring systems. Recent advances in VLMs have significantly improved the performance of VideoQA systems. Models such as \cite{lin-videollava, Vila-wang, Maaz2023VideoChatGPT, clipbert, zhang-videollama, zhang2024llavanextvideo} combine pretrained image/video encoders with large language models to answer questions grounded in visual inputs. However, due to memory and context-length constraints, these models rely on selecting a fixed number of frames from each video before performing inference. The choice of frames is thus critical, as it determines the amount and quality of information the model receives.\par
The dominant approach to frame selection in current VideoQA systems is uniform sampling, where frames are selected at equidistant intervals throughout the video. For instance, Video-LLaVA\cite{lin-videollava} adopts uniform frame sampling as its default strategy to generate a broad, shallow representation of video content. CLIPBERT\cite{clipbert}, similarly, uses either uniform or random sampling prior to multi-modal fusion. While these methods are simple and computationally efficient, they are task-agnostic and often fail to capture semantically relevant frames or exclude rare but crucial visual events needed to answer specific questions. Moreover, uniform strategies tend to introduce redundant information, leading to inefficiencies and poor performance on temporally complex queries.\par
A few recent efforts like \cite{Vila-wang, tang2025adaptivekeyframesamplinglong} attempt to improve upon uniform sampling using learned or heuristic selection mechanisms. However, such methods typically require additional supervision, end-to-end retraining, or are tightly coupled to specific model architectures, making them hard to generalize across tasks or models. Despite these challenges, frame selection remains a largely overlooked bottleneck in modern VideoQA pipelines.\par
To address this, we propose a training-free, model-agnostic frame selection strategy based on SMI functions. These functions provide a principled way to select subsets that are both representative and diverse, with theoretical guarantees for near-optimality under greedy selection \cite{Harshaw2021}. While submodular optimization has been extensively applied in video summarization, it has not yet been adopted for frame selection in VideoQA, where query relevance is paramount. \par To our knowledge, this is the first application of submodular frame selection for large VLMs in VideoQA, establishing a new paradigm for resource-efficient and accurate multi-modal AI. In this paper, we introduce a submodular frame selection pipeline using the SUBMODLIB framework\cite{kaushal-submodlib}, which selects a subset of frames that jointly optimize for temporal diversity and query relevance. Our approach is fully compatible with existing VideoQA models like Video-LLaVA\cite{lin-videollava} and can be seamlessly integrated into the inference pipeline by replacing uniform sampling, without modifying the model architecture or requiring retraining. Experiments across the MVBench dataset demonstrates that query-based frame selection significantly improves answer accuracy and interpretability. Our contributions include:
\begin{itemize}
    \item We propose query-based frame selection for VideoQA models such as Video-LLaVA \cite{lin-videollava} and LLaVA-NeXT \cite{zhang2024llavanextvideo}, showcasing the choice of diverse and representative frame subsets tailored to the query.
    \item Extensive experiments show that our submodular selection method matches or improves accuracy over uniform sampling, demonstrating both performance gains and resource-efficient inference on standard VideoQA benchmarks.
\end{itemize}
\section{Related Work}
\subsection{Submodular Optimization in Video Summarization}
Submodular optimization has in recent times been leveraged for selecting representative subsets from large video collections, particularly in the context of video summarization \cite{Kaushal2020,Kaushal2019}. The core strength of submodular functions lies in their diminishing returns property, which ensures both diversity and relevance in the selected subsets. Sahoo et al. \cite{Sahoo2017} propose a unified multi-faceted video summarization framework that employs submodular optimization to generate key-frames, video skims, and entity-based summaries, supporting both extractive and query-focused settings for interactive summarization of long videos. Works such as \cite{NIPS2014_5d3b9e06} introduced submodular models to create visually diverse summaries, while \cite{7298928} utilized facility location and coverage functions to capture the most informative frames. Mirzasoleiman et al. \cite{Mirzasoleiman2017} address personalized video summarization in streaming environments by introducing streaming local search, a single-pass algorithm with constant approximation guarantees under various constraints, enabling real-time processing of massive video streams. Feldman et al.\cite{Feldman2018} extend this direction with Sample-Streaming, a one-pass submodular maximization method that strategically sub-samples data to achieve tight approximation ratios for both monotone and non-monotone objectives while reducing memory and computation costs. These techniques have shown significant success in reducing redundancy and maintaining the semantic essence of videos. However, despite their proven utility in summarization, submodular methods have not been widely explored in video question answering, where informative and question-relevant frame selection is equally, if not more, critical. This gap motivates our exploration of submodular optimization for frame subset selection in VideoQA settings.

\subsection{Vision-Language Models for Video Question Answering}
Recent advances in VideoQA have been primarily driven by the rise of VLMs that extend large language models to multimodal video inputs. Models such as \cite{lin-videollava, zhang-videollama, Maaz2023VideoChatGPT, Vila-wang, li2023blip2bootstrappinglanguageimagepretraining} integrate video frame encoders with transformer-based language decoders to produce responses grounded in video content. In addition, several vision-language frameworks have been proposed across both image and video domains. CLIPBERT \cite{lei2021clipbert} pioneered an end-to-end sparse sampling approach for joint video-text modeling. BLIP-2 \cite{li2023blip2} connects frozen image encoders with large language models to enable parameter-efficient multimodal learning. Video-LLaVA \cite{lin2023videollava} unifies image and video feature processing before LLM integration, while Video-ChatGPT \cite{Maaz2023VideoChatGPT} fine-tunes a video-adapted visual encoder with an instruction-tuned LLM. ViLA \cite{wang2023vila} introduces a question-aware frame prompter for more targeted temporal reasoning, and Video-LLaMA \cite{zhang2023videollama} incorporates both visual and audio modalities for richer comprehension.\par
Foundational multimodal models include ALBEF \cite{li2021albef} (contrastive alignment before fusion), Flamingo \cite{alayrac2022flamingo} (few-shot VLM for multimodal tasks), PaLI \cite{chen2023pali} (scaling multimodal transformers to 55B parameters), and MiniGPT-4 \cite{zhu2023minigpt4} (bridging pre-trained vision encoders with LLaMA for conversational image understanding). Earlier video-language pretraining approaches include HERO \cite{li2020hero}, MERLOT \cite{zellers2021merlot}, and Frozen in Time \cite{bain2021frozenintime}, which laid the groundwork for modern VideoQA architectures.\par
\subsection{Key frame sampling techniques}
These models typically rely on fixed sampling strategies, either uniform or random sampling to extract a limited number of frames per video, dictated by memory and context length constraints. Although effective to an extent, such sampling methods often result in shallow temporal understanding and can miss question-relevant frames. Recent advances in video understanding have recognized these limitations and proposed various learnable frame selection mechanisms to address this challenge. Works like KeyVideoLLM \cite{Liang2024} introduce text-video frame similarity-based key-frame selection that achieves remarkable data compression rates while maintaining high performance in video question-answering tasks. Similarly, the ShareGPT4Video \cite{Chen2024} dataset demonstrates the importance of semantic-aware key-frame extraction, utilizing CLIP-based similarity measures to maintain temporal coherence while reducing redundancy. Han et al. \cite{Han2024} propose self-adaptive sampling strategies for efficient VideoQA, introducing Most Implied Frames (MIF) and Most Dominant Frames (MDF) methods that leverage pretrained image-text models to identify question-relevant frames without additional training. Some recent works like \cite{Vila-wang} introduce learnable frame selection mechanisms, but they do not leverage principled subset selection methods like submodular optimization. Our work proposes integrating submodular-based frame selection as a pre-processing step to improve context construction for VLM-based VideoQA models. \par

\section{Methodology}
To accomplish the purpose of query-based selection of frames, we chose Submodular functions as they provide a principled, efficient, and effective way to select informative, diverse, and non-redundant subset of video frames. We have used the SUBMODLIB library developed by Kaushal et.al.\cite{kaushal-submodlib}. We provide a brief description of the functions in this section :
\subsection{Submodular functions}
Submodular functions are a special class of set functions that exhibit a diminishing returns property, making them highly suitable for a range of optimization tasks in machine learning, information retrieval, and combinatorial optimization. For a set function, 
\begin{equation}
     f : 2^{\Omega} \rightarrow \mathbb{R}
\end{equation}
 Submodularity intuitively means that the incremental gain from adding an element to a subset decreases as the subset grows. Submodular functions model notions like coverage, diversity, and representativeness, which are crucial in data subset selection, summarization, active learning, and more. Some of the key Types of Submodular Functions under these three categories are:
\begin{enumerate}
    \item [1.] Coverage: Model representativeness by maximizing similarity to a representative subset.(Set cover function, Feature Based function)
    \item [2.] Representativeness: Capture coverage by maximizing the number of unique concepts covered by the subset. (Facility Location function, Graph Cut function)
    \item [3.] Diversity: Encodes division or separation in graphs, useful in community detection and segmentation. (Disparity function, Determinantal Point Processes function )
\end{enumerate}
\subsection{Submodular Mutual Information Functions}
Submodular Mutual Information (SMI) functions are designed to select frames from a video that are not only diverse and representative, but also highly relevant to a given text query. In this context, let set \textbf{A} represent the query and set \textbf{B} represent the video frames. The SMI function quantifies the mutual information between these two sets, enabling the selection of frames from \textbf{B} that best align with the content of \textbf{A} while maintaining diversity and coverage. It is formally defined as:
\begin{equation}
     I_f(A;B) = f(A) + f(B) - f(A \cup B)
\end{equation}
This measures the similarity between the query set \textbf{B} and the video frame set \textbf{A}. It is important to ensure that both textual and visual inputs are represented in the same feature space for the function to effectively identify the most relevant frames. 
\begin{figure*}
\centering
\includegraphics[width =\textwidth]{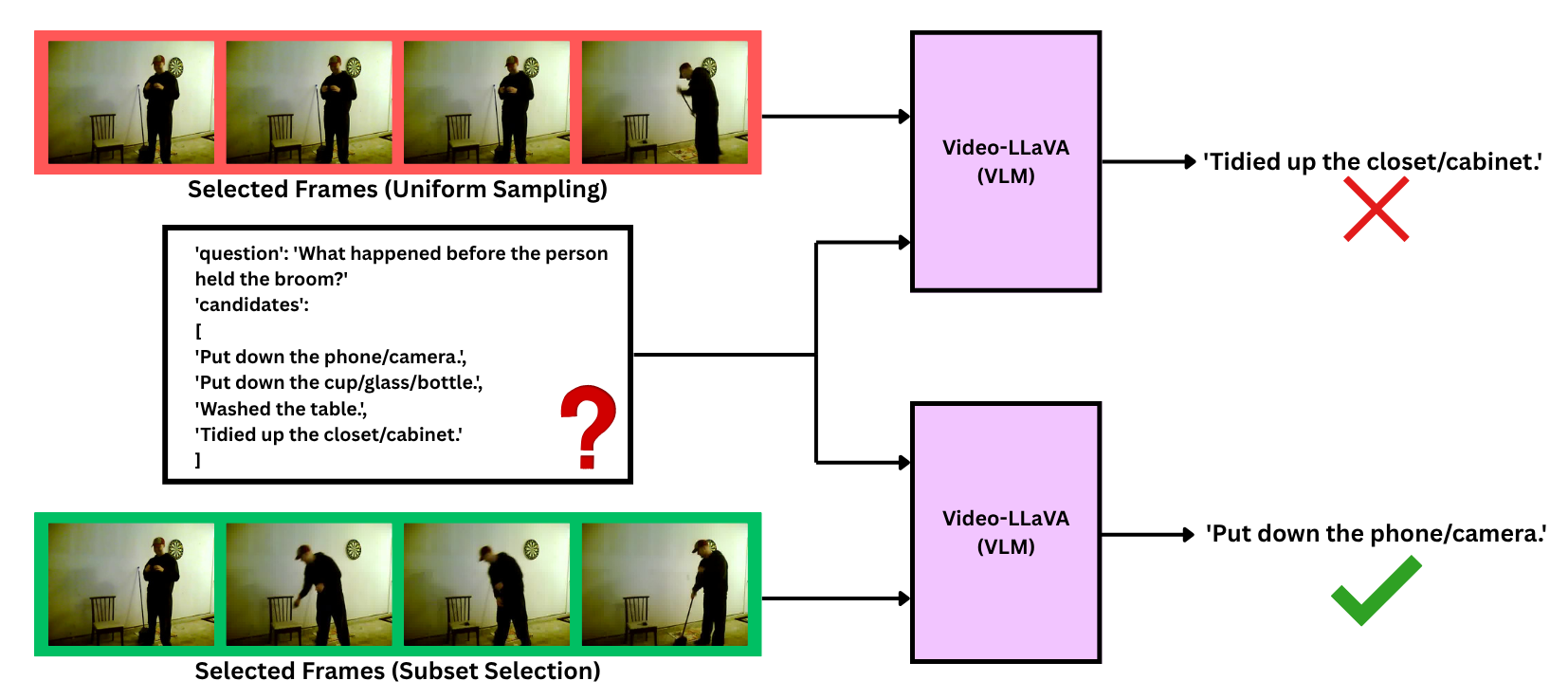}
\caption{Figure illustrates uniform sampling and subset selection passed to VideoLLaVA\cite{lin-videollava}}
\label{fig2}
\end{figure*}
In this study, we considered two SMI functions namely facility location mutual information (FLMI) and graph cut mutual information (GCMI) functions which model query relevance and diversity to different extents.
\begin{enumerate}
    \item Facility Location Mutual Information (FLMI): The Facility Location function aims to capture representativeness by selecting a subset of items that effectively serves as a representative set, similar to the role of centroids or medoids in clustering. FLMI is a specific formulation of the facility location function and is expressed as:
\begin{equation}
     I_f(A;B) =  \sum_{i \in V} min (\max_{j \in A} s_{ij} , \eta \max_{j \in B} s_{ij} )
\end{equation}
Where V is the target set or subset selected and $s_{ij}$ denotes the similarity between data points $i$ and $j$. FLMI often reaches a saturation point, meaning that once the query requirements are met, it gains no additional benefit from selecting another data point related to the query. Moreover, whereas GCMI strongly prioritizes query relevance, FLMI stands at the opposite extreme, emphasizing diversity and comprehensive query coverage rather than strict relevance.
    \item Graph Cut Mutual Information (GCMI) : GCMI uses an undirected weighted graph with nodes as frame embeddings and the edges as similarity between the frame embeddings . GCMI lies at one end of the spectrum favoring only query-relevance and is defined as:
    \begin{equation}
         I_f(A;B) = 2\lambda\sum_{i \in A}\sum_{j \in B} s_{ij}
    \end{equation}
     Where $\lambda$ governs the tradeoff between representation and diversity. When $\lambda$
 becomes large, the graph cut function also tries to model diversity in the subset. In our study we have taken $\lambda$ as 1 and hence focus more on query relevance. 
\end{enumerate}

\section{Implementation Details}
The procedure for frame selection by SMI functions will be discussed in this section. Also, the parameters and other details of the two VLMs, Video-LLaVA and LLaVA-NeXT will be discussed.
\subsection{Frame and Text embeddings:}
A video $V$ with $N$ uniformly or strategically sampled frames can be expressed as:  
\[
V = \{F_1, F_2, \dots, F_N\}
\]  

Each frame $F_i$ is passed through the \textbf{CLIP} image encoder (ViT-B/32 in our case), which first divides the image into patches, linearly projects them, and processes them through a transformer to produce a $d$-dimensional embedding vector:  
\[
\mathbf{f}_i = \mathrm{CLIP}_{\text{img}}(F_i), \quad \mathbf{f}_i \in \mathbb{R}^d
\]  

Similarly, the textual query $T$ is tokenized (note that the query could a single one or even multiple queries, SMI functions can handle multiple queries) and passed through the CLIP text encoder, which uses a transformer to map it into the same $d$-dimensional joint embedding space:  
\[
\mathbf{t} = \mathrm{CLIP}_{\text{text}}(T), \quad \mathbf{t} \in \mathbb{R}^d
\]  

This yields a sequence of frame embeddings:  
\[
E_V = \{\mathbf{f}_1, \mathbf{f}_2, \dots, \mathbf{f}_N\}
\]  

and a single text embedding $\mathbf{t}$, both lying in the same multimodal space. This shared space enables direct computation of similarity (e.g., cosine similarity) between the frames and the query, allowing alignment and selection based on semantic relevance.
To apply these functions effectively, both the query (typically a text description or question) and the video frames must be represented in the same feature space, which is crucial for meaningful similarity computation. To achieve this alignment, we employed the CLIP model \cite{Radford2021}, a powerful multimodal model that can generate embeddings from both images and text in a unified feature space. This model, trained jointly on image-text pairs, enables us to embed video frames (as images) and user queries (as text) into comparable vectors, facilitating the computation of mutual information between them. 
\subsection{Query-based frame selection using SMI functions:}
 As discussed in the previous section, we have employed the SUBMODLIB library and for our use case, we specifically focused on the query-based mutual information functions. These mutual information-based functions are designed to select the most informative and representative subset of data (in our case, video frames) with respect to a given query. The core idea is to measure the amount of mutual information shared between the query and each candidate item (frame), ensuring that the selected frames are both relevant and diverse. Figure \ref{fig1} illustrates this embedding process, where CLIP is used to encode both modalities into a shared representation space. These embeddings are then fed into SMI functions to select a subset of video frames that best capture the information content relevant to the input query.
\subsection{VideoQA using VLMs: Video-LLaVA and LLaVA-NeXT}
\subsubsection{Training Setup}
Our experiments employ two large VLMs namely Video-LLaVA-7B and LLaVA-NeXT-Video as the base architectures. Fine-tuning is performed using the QLoRA approach, which enables efficient low-bit quantized training without sacrificing model performance. To balance computational efficiency and generalization, we update only the aided blocks components specifically designed to enhance video–text alignment while keeping the remaining parameters frozen. This selective training strategy significantly reduces computational overhead and mitigates overfitting. \par
For evaluation, each sample consists of a short video accompanied by a textual prompt containing a question and five multiple-choice options (A–E) in a fixed format. After inference, the model’s output undergoes post-processing to remove extraneous spaces, and the first valid choice (A–E) is extracted as the final predicted answer.
\subsubsection{Video-LLaVA implementation:}
Video-LLaVA is built upon Vicuna v1.5, which itself is based on the LLaMA architecture. The model employs LanguageBind encoders, initialized from OpenCLIP, to align both image and video features into a unified visual feature space. It processes videos by uniformly sampling a fixed number of frames (typically eight), without a designated temporal-pooling module. The shared projection layer then combines these aligned representations with language input, enabling efficient fusion of visual and textual modalities and achieving strong video–text alignment and understanding.
\subsubsection{LLaVA-NeXT Implementation:}
LLaVA-NeXT-Video leverages the CLIP ViT-L/14@336 vision encoder, offering enhanced image and video comprehension capabilities compared to its predecessors. It employs the AnyRes technique to represent each video frame as multiple high-resolution patches, enabling the model to retain fine-grained details. For multimodal fusion, it incorporates both temporal attention and cross-attention mechanisms, which facilitate deeper integration of video features with textual inputs.\par
The architecture supports zero-shot transfer from image-based to video-based understanding, accommodates long video sequences, and incorporates Direct Preference Optimization (DPO) to further refine performance. These enhancements collectively enable LLaVA-NeXT-Video to handle a wider range of video–language tasks with improved accuracy and robustness.

\subsection{VideoQA Dataset}
We evaluate our submodular query-based frame selection approach using the MVBench dataset \cite{MVBench}, a comprehensive Multi-modal Video understanding benchmark comprising 20 diverse and temporally complex tasks. The dataset consists a total of 4000 videos each at 30 fps divided equally for every task category, and we use an 80/20 split for training and testing. The average clip length for each task category can be found in Table \ref{tab:avg_video_lengths}. MVBench was specifically designed to test the dynamic reasoning capabilities of VLMs by converting static image tasks into temporally-aware video tasks. It covers a wide range of challenges, including Action Sequence (AS), Action Prediction (AP), Action Antonym (AA), Fine-grained Action (FA), Unexpected Action (UA), Object Existence (OE), Object Interaction (OI), Object Shuffle (OS), Moving Direction (MD), Action Localization (AL), Scene Transition (ST), Action Count (AC), Moving Count (MC), Moving Attribute (MA), State Change (SC), Fine-grained Pose (FP), Character Order (CO), Egocentric Navigation (EN), Episodic Reasoning (ER), and Counterfactual Inference (CI). These tasks collectively enable a robust evaluation of video question answering models beyond static spatial understanding, making MVBench an ideal benchmark for our work.

\section{Results}

Table \ref{table1} presents the accuracy scores for the two VLM models on the MVBench dataset, averaged over 20 actions. The results indicate that both FLMI and GCMI-based frame selection outperform uniform selection for both frame counts, 4 and 12. Due to memory constraints when using an NVIDIA A6000 48 GB RAM GPU, we were limited to processing a maximum of 14 frames. Consequently, we chose 12 frames as the upper limit and 4 frames as the lower limit. This range is reasonable given that the video lengths span from 5 to 30 seconds. A detailed analysis of FLMI’s performance across the 20 actions for the two VLMs is provided in the following section. While GCMI tends to prioritize frames that are more directly relevant to the query, FLMI achieves a balance between diversity and relevance. Based on the results, GCMI demonstrates superior performance compared to FLMI, suggesting it is more effective at capturing query-relevant frames.\par
In contrast, AKS under performs relative to both mutual information–based methods and even the uniform baseline. This may be due to the choice of parameters we selected which are mean deviation threshold, standard deviation threshold, and depth. They were set to 0.1, –200, and 5, respectively, instead of the default values in the original repository. When the default AKS parameters were applied to the MVBench dataset, the method failed to select any frames. The processing speed of AKS was measured at 12.5 fps, slightly lower than the 13 fps achieved by FLMI subset selection.\par
As noted in the introduction, query-based frame selection using submodular functions is well-suited for video frame selection due to its diminishing returns property, which naturally reduces redundancy and ensures diverse, informative subsets. Such functions enable efficient, near-optimal greedy algorithms with theoretical guarantees and can be adapted to various objectives, including maximizing relevance, coverage, or diversity. Empirically, submodular methods consistently outperform similarity-based approaches in tasks such as video summarization, VideoQA preprocessing, and surveillance frame selection. From our experiments also, we observe that submodular frame selection outperforms heuristic-based methods, which may require further fine-tuning depending on the dataset. Our approach performs particularly well on questions that are less dependent on temporal information, such as object existence, spatial localization, and attribute identification.
\begin{table}[h]
    \centering
    \setlength{\tabcolsep}{6pt}
    \renewcommand{\arraystretch}{1.4}
    \resizebox{\linewidth}{!}{
        \begin{tabular}{lccccc}
            \toprule
            \textbf{Method} & \textbf{\#Frames} & \textbf{Uniform} & \textbf{AKS~\cite{tang2025adaptivekeyframesamplinglong}} & \textbf{FLMI} & \textbf{GCMI}\\
            \midrule
            Video-LLaVA~\cite{lin-videollava} 
                & \makecell{4 \\ 12} & \makecell{33.21\\32.38} &\makecell{32.62 \\ 33.40} & \makecell{35.13\\36.08} & \makecell{37.10 \\ 36.45} \\
            \Xhline{0.5pt}
            LLaVA-NeXT~\cite{zhang2024llavanextvideo} 
                & \makecell{4 \\ 12} & \makecell{31.48\\32.82}  & \makecell{34.84 \\ 33.28}& \makecell{34.32 \\ 34.11} & \makecell{\textbf{37.22} \\ \underline{36.70}} \\
            \bottomrule
        \end{tabular}
    }
    \caption{\textbf{Average video question answering accuracy (\%) of Video-LLaVA and LLaVA-NeXT on the MVBench~\cite{MVBench} dataset, evaluated across different frame selection strategies and varying numbers of selected frames. Facility Location (FLMI) and Graph Cut (GCMI) refer to mutual information-based subset selection functions from SUBMODLIB~\cite{kaushal-submodlib}.The best result is highlighted in bold, and the second best is underlined}}
    \label{table1}
    \vspace{-0.4 cm}
\end{table}

\section{Quantitative Analysis on the dataset}
In this section, we analyse quantitatively the accuracy values obtained using the FLMI function for the two VLMs for all the 20 video actions.Table \ref{table 2} details the accuracy values of two VLMs, Video-LLaVA and LLaVA-NeXT, on the MVBench dataset when using FLMI function for frame selection. The results are presented for both 4 and 12 selected frames, allowing for a qualitative analysis of FLMI's impact on individual temporal understanding tasks.
\subsection{Overall Performance with FLMI} While the overall average accuracy for both Video-LLaVA and LLaVA-NeXT improves with the implementation of FLMI, indicating its general effectiveness as a query-based frame selection strategy as noted in the abstract, a detailed look at individual tasks reveals a varied impact. For Video-LLaVA, the average accuracy increased from 33.21\% to 35.13\% with 4 frames and from 32.38\% to 36.08\% with 12 frames. Similarly, for LLaVA-NeXT, the average accuracy improved from 31.48\% to 34.32\% with 4 frames and from 32.82\% to 34.11\% with 12 frames.

\begin{table*}[t!]
    \centering
    \setlength\tabcolsep{1pt}
    \resizebox{1.0\textwidth}{!}{
        \begin{tabular}{l|l|c|c|c|c|c|c|c|c|c|c|c|c|c|c|c|c|c|c|c|c|c|c}
        \hline\hline
        \textbf{Model} & \textbf{LLM} & \textbf{\#Frames} & \textbf{Avg} & \textbf{AS} & \textbf{AP} & \textbf{AA} & \textbf{UA} & \textbf{OE} & \textbf{OI} & \textbf{OS} & \textbf{MD} & \textbf{AL} & \textbf{ST} & \textbf{AC} & \textbf{MC} & \textbf{MA} & \textbf{SC} & \textbf{CO} & \textbf{EN} & \textbf{CI} &\textbf{FA} &\textbf{FP} &\textbf{ER}\\
        \hline\hline
       
        Video-LLaVA~\cite{lin-videollava} & Vicuna-7B & \makecell{4\\12} 
        & \makecell{33.21 \\ 32.38}
        & \makecell{\textbf{30} \\ 23.84} 
        & \makecell{27.5 \\ 20} 
        & \makecell{35 \\ 47.5}
        & \makecell{21.62 \\ 28.64} 
        & \makecell{52.5 \\ 40} 
        & \makecell{30 \\ 40} 
        & \makecell{35 \\ \underline{40}} 
        & \makecell{37.5 \\ 35} 
        & \makecell{30 \\ 25} 
        & \makecell{30 \\ 32.5} 
        & \makecell{32.5 \\ 32.5}
        & \makecell{17.5 \\ 20} 
        & \makecell{\underline{47.5} \\ 32.5} 
        & \makecell{32.5 \\ 38.9} 
        & \makecell{\underline{42.5}\\35} 
        & \makecell{\underline{34.21} \\ 30.68} 
        & \makecell{30 \\ 40} 
        & \makecell{29.4 \\ 28.1}
        &\makecell{25 \\ 27.5} &\makecell{21.6 \\ 20}\\
        \hline
        Video-LLaVA w/ FLMI & Vicuna-7B & \makecell{4\\12} 
        & \makecell{35.13 \\ 36.08} 
        & \makecell{27.5\\27.64}
        & \makecell{30\\27.5} 
        & \makecell{57.5\\52.5}
        & \makecell{28.32\\36.84} 
        & \makecell{50\\42.5} 
        & \makecell{32\\42} 
        & \makecell{29.5\\\textbf{47.5}} 
        & \makecell{40\\\textbf{47.5}} 
        & \makecell{28.5\\27.5} 
        & \makecell{32.5\\40} 
        & \makecell{39.5\\\underline{41.5}}
        & \makecell{27.5\\22.5} 
        & \makecell{\textbf{55}\\\underline{47.5}} 
        & \makecell{37.5\\43.58} 
        & \makecell{40\\32.5} 
        & \makecell{28.94\\25.05} 
        & \makecell{\underline{42.5}\\37.5} 
        & \makecell{23.5 \\ 30} 
        & \makecell{28.5\\25} 
        & \makecell{20\\25}\\
        \hline
        
        Video-LLaVA w/ GCMI & Vicuna-7B & \makecell{4\\12} 
        & \makecell{37.10\\36.45} 
        & \makecell{25.5\\21.51} 
        & \makecell{\underline{40}\\35} 
        & \makecell{\textbf{75}\\50.37} 
        & \makecell{\underline{45.24}\\36.21} 
        & \makecell{55\\52.5} 
        & \makecell{35\\50} 
        & \makecell{\underline{40}\\\textbf{47.5}} 
        & \makecell{35\\32.5} 
        & \makecell{20.5\\15.5} 
        & \makecell{\textbf{47.5}\\35} 
        & \makecell{25.5\\30.02} 
        & \makecell{\textbf{32.59}\\25.78} 
        & \makecell{42.5\\35.89} 
        & \makecell{50\\ \underline{53.84}} 
        & \makecell{32.5\\32.5} 
        & \makecell{27.68\\ \underline{34.21}} 
        & \makecell{32.5\\\textbf{47.5}} 
        & \makecell{23.52\\25} 
        & \makecell{\underline{35} \\ \underline{35}} 
        & \makecell{21.9\\33.33}\\
\hline\hline

        \hline
        LLaVA-NeXT~\cite{zhang2024llavanextvideo} & Vicuna-7B & \makecell{4\\12} 
        & \makecell{31.48 \\ 32.82}
        & \makecell{27.5\\ \underline{28.94}} 
        & \makecell{32.5\\27.7} 
        & \makecell{\underline{65} \\ 50} 
        & \makecell{24.32 \\ 33.11} 
        & \makecell{\textbf{63.63} \\ 52.5} 
        & \makecell{\underline{42.5} \\ 32.21} 
        & \makecell{\underline{40}\\ 35.46} 
        & \makecell{23.52\\17.5} 
        & \makecell{32.5\\28.94} 
        & \makecell{30 \\ 32.5} 
        & \makecell{22.5 \\\textbf{45.38}}
        & \makecell{17.14 \\ 25} 
        & \makecell{32.35 \\ 35} 
        & \makecell{35 \\ 50} 
        & \makecell{25 \\ 37.47}
        & \makecell{26.31 \\ 32.21} 
        & \makecell{22.85 \\ 22.5} 
        & \makecell{20.6 \\ 20} 
        & \makecell{27.5 \\ 25} 
        & \makecell{18.9 \\ 25}\\
        \hline
        
        LLaVA-NeXT w/ FLMI & Vicuna-7B & \makecell{4\\12} 
        & \makecell{34.32 \\ 34.11}
        & \makecell{27.5\\17.89} 
        & \makecell{32.5\\\textbf{46.15}} 
        & \makecell{42.5\\50} 
        & \makecell{40.54\\38.88} 
        & \makecell{57.57\\47.5} 
        & \makecell{35\\23.07} 
        & \makecell{29.5\\23.07} 
        & \makecell{25.52\\30} 
        & \makecell{\underline{35}\\31.22} 
        & \makecell{29.5\\30} 
        & \makecell{27\\29.03}
        & \makecell{25.71\\27.5} 
        & \makecell{35.29\\32.5} 
        & \makecell{46\\\textbf{58.33}} 
        & \makecell{38.5\\ 42.10}
        & \makecell{29.94\\28.94} 
        & \makecell{40\\25} 
        & \makecell{21 \\ \textbf{37.5}} 
        & \makecell{30 \\ 30} 
        & \makecell{\underline{37.83} \\ 33}\\
        \hline
        
        LLAVA-NeXT w/ GCMI & Vicuna-7B & \makecell{4\\12}
        & \makecell{\textbf{37.22} \\ \underline{36.70}}
        & \makecell{27\\35.89} 
        & \makecell{29.57\\33.52}
        & \makecell{52.5\\50.25} 
        & \makecell{\textbf{46.57}\\39.47}
        & \makecell{\underline{59.47}\\47.5} 
        & \makecell{35.55\\\textbf{42.57}} 
        & \makecell{31.5\\29.07} 
        & \makecell{\underline{42.5}\\30.87} 
        & \makecell{30.78\\\textbf{38.57}} 
        & \makecell{37.5\\ \underline{42.57}} 
        & \makecell{37\\32.72} 
        & \makecell{28.5\\ \underline{32.5}} 
        & \makecell{37.5\\35.47} 
        & \makecell{47.23\\35.89} 
        & \makecell{30.5\\\textbf{43.57}} 
        & \makecell{33.31\\\textbf{38.84}} 
        & \makecell{35\\32.57} 
        & \makecell{\underline{29.47}\\25.57} 
        & \makecell{\underline{35}\\\textbf{35.57}} 
        & \makecell{\textbf{38.13}\\30.77}\\
\hline

        \end{tabular}
    }
    \caption{
        VideoQA accuracy (\%) on the MVBench dataset\cite{MVBench} under different tasks for Video-LLaVA and LLaVA-NeXT using 4 and 12 frames using FLMI function. 18 of the 20 tasks perform better with FLMI subset selection as compared to uniform sampling for 12 frames. The description of each task can be found in Section 4.2. The average video length of each task can be found in Table \ref{tab:avg_video_lengths}. The best result for each task is highlighted in bold, and the second best is underlined.
    }
    \label{table 2}
\end{table*}

\begin{table}[t!]
\centering
\setlength\tabcolsep{6pt}
\begin{tabular}{l|c}
\hline\hline
\textbf{Task} & \textbf{Avg. Video Length (s)} \\
\hline\hline
Action Sequence (AS) & 44.32 \\
Action Prediction (AP) & 36.77 \\
Action Antonym (AA) & 5.17 \\
Unexpected Action (UA) & 6.30 \\
Object Existence (OE) & 5.12 \\
Object Interaction (OI) & 19.92 \\
Object Shuffle (OS) & 28.79 \\
Moving Direction (MD) & 5.12 \\
Action Localization (AL) & 30.66 \\
Scene Transition (ST) & 20.00 \\
Action Count (AC) & 15.38 \\
Moving Count (MC) & 5.12 \\
Moving Attribute (MA) & 5.12 \\
State Change (SC) & 35.00 \\
Character Order (CO) & 29.27 \\
Egocentric Navigation (EN) & 8.67 \\
Counterfactual Inference (CI) & 5.12 \\
Fine-grained Action (FA) & 3.00 \\
Fine-grained Pose (FP) & 2.50 \\
Episodic Reasoning (ER) & 8.97 \\
\hline\hline
\end{tabular}
\caption{
Average video lengths for each video question answering task under MVBench Dataset.
}
\label{tab:avg_video_lengths}
\end{table}

\subsection{Analysis for Video-LLaVA using FLMI}
\subsubsection{Significant Improvements:}
\begin{itemize}
    \item With 4 frames, FLMI notably boosted performance in Action Antonym (AA), increasing accuracy from 35\% to 57.5\%. Moving Count (MC) saw a substantial rise from 17.5\% to 27.5\%, and State Change (SC) experienced a significant leap from 32.5\% to 37.5\%. Scene Transition (ST) also improved from 30\% to 32.5\%. 
    \item With 12 frames, Action Localization (AL) showed a considerable improvement, from 32.5\% to 41.5\%. Moving Count (MC) again demonstrated strong gains, moving from 32.5\% to 47.5\%. Fine-grained Action (FA) also improved from 40\% to 42.5\%.
\end{itemize}
In summary, the datasets AP, AA, UA, OI, ST, MD, AL, MC, MA, and SC showed improved accuracy in both cases. We note that this holds true even when the average video length is 36.77 s. As previously mentioned, frame selection is influenced by the encoder used and in our case it is CLIP encoder. From our study, we observe that in scenarios where capturing global semantic meaning, understanding objects, attributes, and spatial relationships in images is important, query-based frame selection enhances VideoQA accuracy.  
\subsubsection{Notable Decreases:}
We observe a slight drop in accuracy for CO, from 42.5 \% to 40 \% with 4 frames and from 35 \% to 32.5 \% with 12 frames. In contrast, EN shows a substantial decrease, from 34.21 \% to 28.94 \% for 4 frames and from 30.68 \% to 25.05 \% for 12 frames. Both actions require temporal understanding. CO depends on the sequence in which letters are placed, while EN relies on direction and temporal comprehension. As noted earlier, the encoder used does not capture changes over time or causal relationships between frames, leading to this accuracy drop.  
\subsubsection {Marginal/Mixed Changes:} Tasks like Action sequence (AS), Fine grained action (FA), Object shuffle (OS), Object Existence (OE), Action Localisation (AL), Fine grained pose (FP), Episodic Reasoning (ER), and Counterfactual Inference (CI) showed a mixed performance depending on the number of frames. In most cases, accuracy decreased with 4 frames but increased with 12 frames, while in some instances the reverse trend was observed. As mentioned earlier, certain cases required temporal understanding, and in others, the short average video length meant that uniform selection could still capture the necessary frames.

\subsection{Analysis for LLaVA-NeXT using FLMI}  
\subsubsection{Significant Improvements:}
We observe a notable accuracy gain for state change (SC), increasing from 35 \% to 46 \% with 4 frames, and from 50 \% to 58.33 \%. Accuracy also improves for AP, AL, UA, SC, CI, CO, FA, FD, EN, ER, MD, and MC. Unlike in Video-LLaVA, we record an accuracy increase for CO, which may be attributed to the model’s improved ability to comprehend temporal sequences.
\subsubsection{Notable Decreases:}
We also observe a significant decrease in accuracy for OE, OI, and OS, with a minor drop for ST. This may be due to two factors: the short video length, where uniform sampling captures more relevant frames, and the image encoder’s limitation in capturing frames that contribute effectively to temporal reasoning. 
\subsubsection{Marginal/Mixed Changes:} Similar to the previous model, certain categories show a mixed trend, where accuracy increases with 4 frames but decreases with 12 frames, and vice versa. This pattern is observed in the cases of AS, AA, AC, MA, and EN, indicating that the optimal number of frames for these tasks may vary depending on how well the selected frames capture the relevant temporal or contextual information. \par

LLaVA-NeXT enhances temporal reasoning compared to Video-LLaVA by training on large-scale multi-image and video instruction datasets, allowing it to better capture frame-to-frame dependencies and event progression. Its visual expert modules and improved alignment layers help maintain temporal coherence, making it more effective for tasks that demand a nuanced understanding of motion and sequence. Consequently, as expected, LLaVA-NeXT outperforms Video-LLaVA in cases requiring temporal comprehension. Notably, while Character Order (CO) accuracy decreased with Video-LLaVA, it improved with LLaVA-NEXT. \par 
In summary, FLMI generally yields an overall accuracy improvement for both Video-LLaVA and LLaVA-NeXT on the MVBench dataset, though its impact varies considerably across the 20 challenging video tasks. It offers clear benefits for tasks that leverage its balance between diversity and relevance in frame selection. However, the success of query-based frame selection in providing complementary and essential visual information is highly dependent on the encoder used to convert frames and images into embeddings. 

\section{Conclusions}
The accuracy of query-based frame selection is influenced by several factors, including the choice of image and text encoder used to generate embeddings from the video frames and queries, the duration of the video, and the specific VLM utilized. Since SMI functions operate on these embeddings, the quality of frame selection is directly tied to the features extracted by the encoder. In our approach, we used the CLIP encoder, which is particularly effective in capturing spatial relationships between images and text. However, CLIP does not extract temporal features, so the selected frames lack temporal context. As a result, for VideoQA tasks that require temporal reasoning, especially in short videos (around 5 seconds), uniform sampling tends to outperform query-based selection. The framework proposed in this work is encoder-agnostic and can be replaced with encoders that capture temporal dependencies to address this limitation. 

\begin{acks}
The authors acknowledge the financial support by the Government of Maharashtra through the Maharashtra Drone Mission project. 
\end{acks}
\bibliographystyle{ACM-Reference-Format}
\bibliography{references}


\end{document}